\documentclass[10pt,twocolumn,letterpaper]{article}

\usepackage{cvpr}
\usepackage{times}
\usepackage{epsfig}
\usepackage{graphicx}
\usepackage{amsmath}
\usepackage{amssymb}
\usepackage{subfigure}

\usepackage[pagebackref=true,breaklinks=true,letterpaper=true,colorlinks,bookmarks=false]{hyperref}

\newcommand*{\USEIMAGES}{}

\cvprfinalcopy %

\def\cvprPaperID{1775} %

\ifcvprfinal\fi

\newcommand\blfootnote[1]{%
  \begingroup
  \renewcommand\thefootnote{}\footnote{#1}%
  \addtocounter{footnote}{-1}%
  \endgroup
}

\begin{document}

\title{Learning to Compare Image Patches via Convolutional Neural Networks}

\author{Sergey Zagoruyko\\
Universite Paris Est, Ecole des Ponts ParisTech\\
{\tt\small sergey.zagoruyko@imagine.enpc.fr}
\and
Nikos Komodakis\\
Universite Paris Est, Ecole des Ponts ParisTech\\
{\tt\small nikos.komodakis@enpc.fr}\\
}

\maketitle
\begin{abstract}
\vspace{-4pt}
In this paper we show how to learn %
directly from image data (\ie, without resorting to manually-designed features) a general similarity function for comparing image patches, which is a task of fundamental importance for many computer vision problems. To encode such a function, we  opt for  a CNN-based model that is trained to account for a wide variety of changes in image appearance. To that end, we explore and study multiple neural network architectures, which are specifically adapted to this task. We show that such an approach can significantly outperform the state-of-the-art  on several problems and benchmark datasets.
\end{abstract}

\vspace{-15pt}
\section{Introduction}
\blfootnote{Source code and trained models are available online at \url{http://imagine.enpc.fr/~zagoruys/deepcompare.html} (work  supported by  EC project FP7-ICT-611145 ROBOSPECT).}
Comparing patches across images is probably one of the most fundamental tasks in computer vision and image analysis. It is often used as a subroutine that plays an important role in a wide variety of vision tasks. These can range from low-level tasks such as  structure from motion, wide baseline matching, building panoramas, and image super-resolution, up to higher-level tasks such as object recognition, image retrieval, and classification of object categories, to mention a few characteristic examples. 

Of course, the problem of deciding if two patches correspond to each other or not is  quite challenging  as there exist far too many factors that affect the final appearance of an image \cite{nowak:hal-00203958}. These can include changes in viewpoint, variations in the overall illumination of a scene, occlusions, shading, differences in camera settings, \etc.
In fact, this need of comparing patches has given rise to the development of many hand-designed feature descriptors over the past years, including SIFT \cite{LoweSift}, that had a huge impact in the computer vision community.
Yet, such manually designed descriptors may be unable
to take into account in an optimal manner all of the aforementioned factors that determine the appearance of a patch. On the other hand, nowadays one can easily gain access to (or even generate using available software) large datasets that contain  patch correspondences between images      \cite{photo_collections}.
 This begs the following question: can we make proper use of such datasets to automatically
learn a similarity function for image patches ?

\ifdefined\USEIMAGES
\begin{figure}
\small
\begin{center}
\includegraphics[scale=0.45]{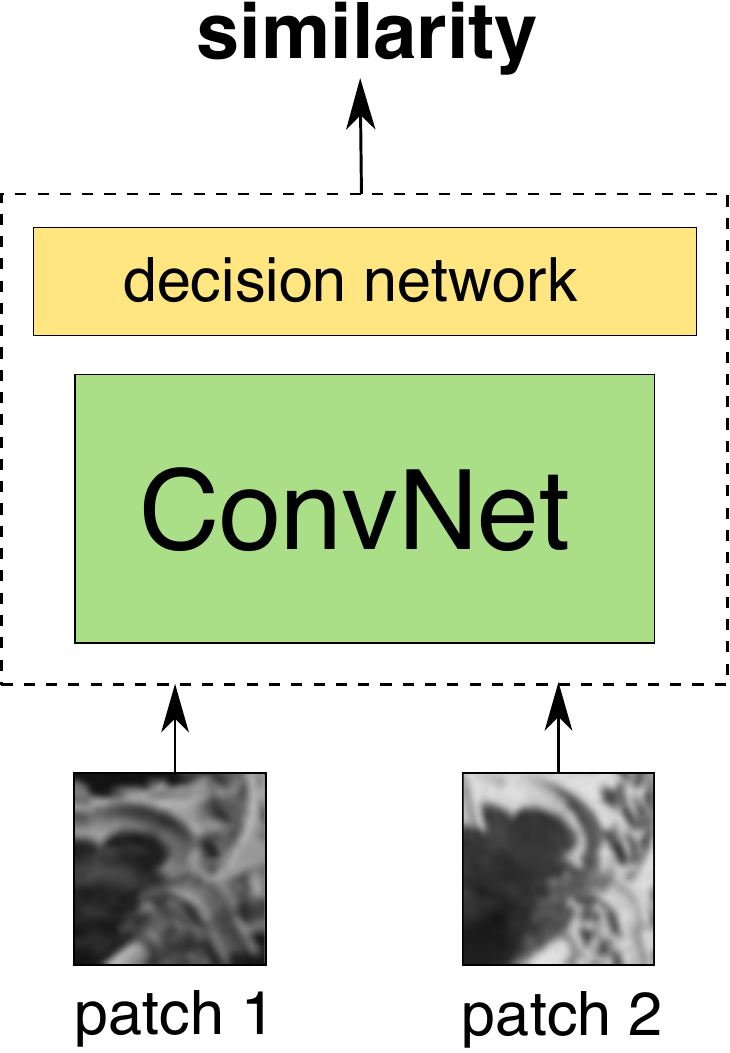}
\end{center}
\vspace{-7pt}
\caption{Our goal is to learn a  general similarity function for image patches. To encode such a function, here we make use of   and explore convolutional neural network architectures.}
\label{fig:intro}
\small
\vspace{-13pt}
\end{figure}
\fi

The goal of this paper is to affirmatively address the above question. Our aim is thus to be able to generate a patch similarity function \emph{from scratch}, \ie, without attempting to use any manually designed features but instead directly learn this function from annotated pairs of  raw image patches. To that end, inspired also by the recent advances in  neural architectures and  deep learning, we choose to represent such a function in terms of a deep convolutional neural network \cite{lecun-88, kriz2012} (Fig.~\ref{fig:intro}). In doing so, we are also interested in addressing the  issue of what   network architecture should  be best used in a task like this. We thus explore and propose various types of networks, having  architectures that  exhibit different trade-offs and advantages.
 In all  cases, to train these networks, we are using as  sole input a large database  that contains pairs of raw image patches (both matching and
non-matching).
This  allows to further improve the performance of our method  simply by enriching this database with more samples 
(as  software for automatically generating such samples is readily available  \cite{photo_tourism}).

To conclude this section, the paper's main contributions  are as follows:
(i) We learn directly from image data (\ie, without any manually-designed features) a  general similarity function  for patches that can implicitly take into account various types of transformations and effects (due to \eg, a wide baseline, illumination, \etc). (ii) We  explore and propose a variety of different neural network models adapted for  representing such a function, highlighting  at the same time network architectures 
that  offer improved  performance. 
  as in \cite{Simonyan14}.
(iii) We apply our approach on several problems and benchmark datasets, showing that it significantly outperforms the state-of-the-art and that it leads to feature descriptors  with much better performance than manually designed descriptors  (\eg, SIFT, DAISY) or     other learnt descriptors
  as in \cite{Simonyan14}. Importantly, due to their convolutional nature, the resulting descriptors are very efficient to compute even in a dense manner.

\section{Related work}

The conventional approach to compare patches is to use descriptors and a squared euclidean distance. Most feature descriptors
are hand-crafted as SIFT \cite{LoweSift} or DAISY \cite{Tola08}. Recently, methods for learning a descriptor have been proposed \cite{Trzcinski12} (\eg, DAISY-like descriptors learn pooling regions and dimensionality reduction \cite{BHW10}). Simonyan \etal 
\cite{Simonyan14} proposed a convex procedure for training on both tasks.

Our approach, however, is inspired by the recent success of convolutional neural networks \cite{Razavian, deepface, Szegedy, depth_map}. Although these models   involve 
a highly non-convex  objective function during training, they have shown outstanding results in various tasks \cite{Razavian}. Fischer \etal \cite{comparison} analysed the performance of convolutional descriptors from AlexNet 
network (that was trained on Imagenet dataset \cite{kriz2012}) on the well-known Mikolajczyk dataset \cite{MS05} 
and showed that these convolutional descriptors outperform SIFT in most cases except blur. They also 
proposed an unsupervised training approach for deriving descriptors that outperform both SIFT and Imagenet trained network.

Zbontar and LeCun in \cite{Zbontar} have recently proposed a CNN-based approach to compare patches 
for computing cost in small baseline stereo problem and shown the best performance in KITTI dataset. 
However, the focus of that work was   only on comparing pairs that consist of very small patches like the ones in narrow baseline stereo. In contrast, here we aim for a   similarity function that can account for a broader set of appearance changes  and can be used in a much wider and more challenging set of applications, including, \eg, wide baseline stereo, 
feature matching and image retrieval.

\section{Architectures}
As already mentioned, the input to the  neural network 
is considered to be a pair of image patches.  Our models do not 
impose any limitations with respect to the number of channels in the input patches, \ie, given a dataset with
colour patches the networks could be trained to further increase performance.
However, to be able to compare 
our approach with state-of-the-art methods on existing datasets, we chose to use only grayscale patches during training. 
Furthermore, with the exception of the SPP model described in section \ref{sec:spp}, in all other cases the  patches given as input to the network are  assumed to have  a fixed size of $64\times64$ (this means that original patches may need to be resized to the above spatial dimensions). 

There are several ways in which patch pairs can be processed by the network and how 
the information sharing can take place in this case. For this reason, we explored and tested a variety of models. We start in section \ref{sec:basic_models}  by describing   the three basic neural network architectures that we studied, \ie, 2-channel, Siamese, Pseudo-siamese (see Fig.~\ref{fig:basic_models}), which offer different trade-offs in terms of speed and 
accuracy (note that, as usually, applied patch-matching techniques imply testing a 
patch against a big number of other patches, and so re-using computed information is always useful). Essentially these architectures stem from the 
different way that each of them attempts to address the following question: when composing a similarity function for comparing image patches, do we   first choose to compute  a descriptor for each patch and then create a similarity on top of these descriptors or do we perhaps choose to skip the part related to the descriptor computation and directly proceed with the  similarity estimation? 

In addition to the above basic models, we also describe in section \ref{sec:extra_models} some extra  variations concerning the  network architecture. %
 These variations, which are not mutually exclusive to each other,  can be used in conjunction with any of the basic models described in section \ref{sec:basic_models}. %
Overall, this  leads to  a variety of  models that is possible to be used for the task of comparing image patches.

\ifdefined\USEIMAGES
\begin{figure}
  \begin{center}
  \includegraphics[scale=0.5]{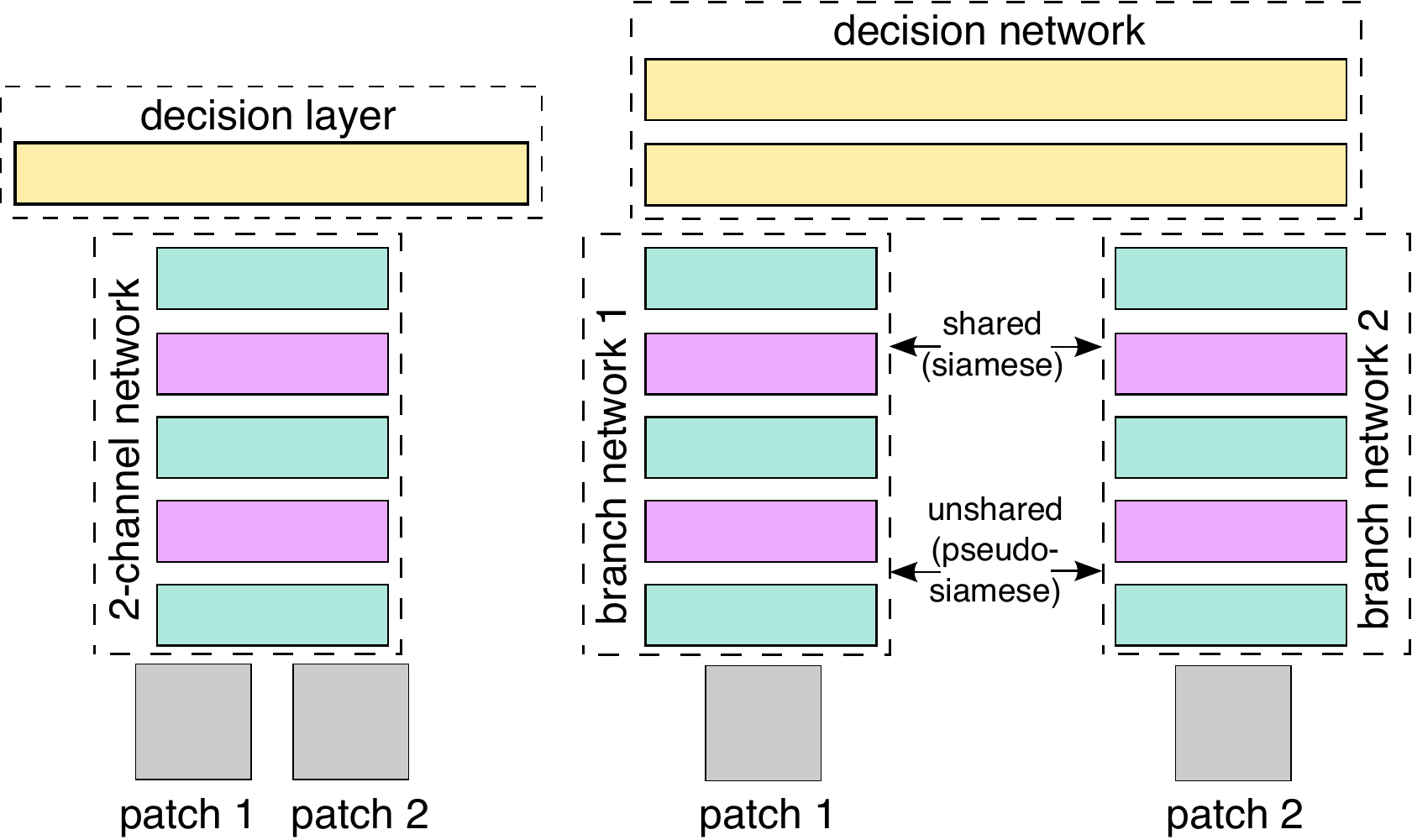}
  \end{center}
  \vspace{-5pt}
  \caption{Three basic network architectures: 2-channel on the left, siamese and pseudo-siamese on the right (the difference between siamese and pseudo-siamese is that the latter does not have shared branches). Color code used: cyan $=$ Conv+ReLU, purple $=$ max pooling, yellow $=$ fully connected layer (ReLU  exists between fully connected layers as well).}
  \label{fig:basic_models}
  \vspace{-5pt}
\end{figure}
\fi

\subsection{Basic models}\label{sec:basic_models}

\textbf{Siamese:} This type of network resembles the idea of having a descriptor \cite{siamese, chopra-05}. There are two branches in the network that share exactly the same architecture and the same set of weights. Each branch takes as input one of the two patches and then applies a series of convolutional, ReLU and max-pooling layers. 
Branch outputs are concatenated and given to 
a top network that consists of linear fully connected and ReLU layers. In our tests we used a top network consisting of 2 linear fully connected layers (each with 512 hidden units) that are separated by a ReLU activation layer.

Branches of the siamese network can be viewed as descriptor computation modules and 
the top network - as a similarity function. For the task of matching two sets of patches at test time, descriptors can first be computed independently using the branches and then matched with the top 
network (or even with a distance function like $l_2$).

\textbf{Pseudo-siamese:} In terms of complexity, this architecture can be considered as being in-between the siamese and  the 2-channel  networks.
More specifically, it has the structure of the siamese net described above except that the weights of the two branches are
 uncoupled, \ie, not shared. This increases the number of parameters that can be adjusted during training and provides more flexibility than
a restricted siamese network, but not as much as the 2-channel network described next. On the other hand, it maintains the efficiency of siamese network at test time.

\textbf{2-channel:} unlike the previous models, here there is no direct notion of descriptor in the  architecture. We simply consider the two patches of  an input pair as a 2-channel image, which is directly fed to the first convolutional layer of the network. In this case, the bottom part of the network  consists of  a series of convolutional, ReLU and max-pooling layers. The output of this part is then given as input to a top module that  consists simply of a fully connected linear decision layer with 1 output.
This network provides greater flexibility compared to the above models as it starts by processing the two patches jointly. Furthermore, it is fast to train, but in general at test time it is more expensive as it requires all 
combinations of patches to be tested against each other in a brute-force manner.

\subsection{Additional models}\label{sec:extra_models}
\textbf{Deep network.} We apply the technique proposed by Simonyan and Zisserman 
in \cite{verydeep}  advising to break up bigger convolutional layers into smaller 3x3 kernels, 
separated by ReLU activations, which is supposed to increase the nonlinearities inside the network and  make the decision function more  discriminative. 
They also report that it might be difficult to initialise such a network, we, however, do not 
observe this behavior and train the network from scratch as usual. In our case,  when applying this technique to our model, the convolutional part of the final architecture turns out to consist of one
convolutional 4x4 layer and 6 convolutional layers with 3x3 layers, separated by ReLU
activations. As we shall also see later in the experimental results, such a change in the network architecture can contribute in further improving performance, which is in accordance with analogous observations made in \cite{verydeep}.

\ifdefined\USEIMAGES
\begin{figure}
\small
\begin{center}
\includegraphics[scale=0.5]{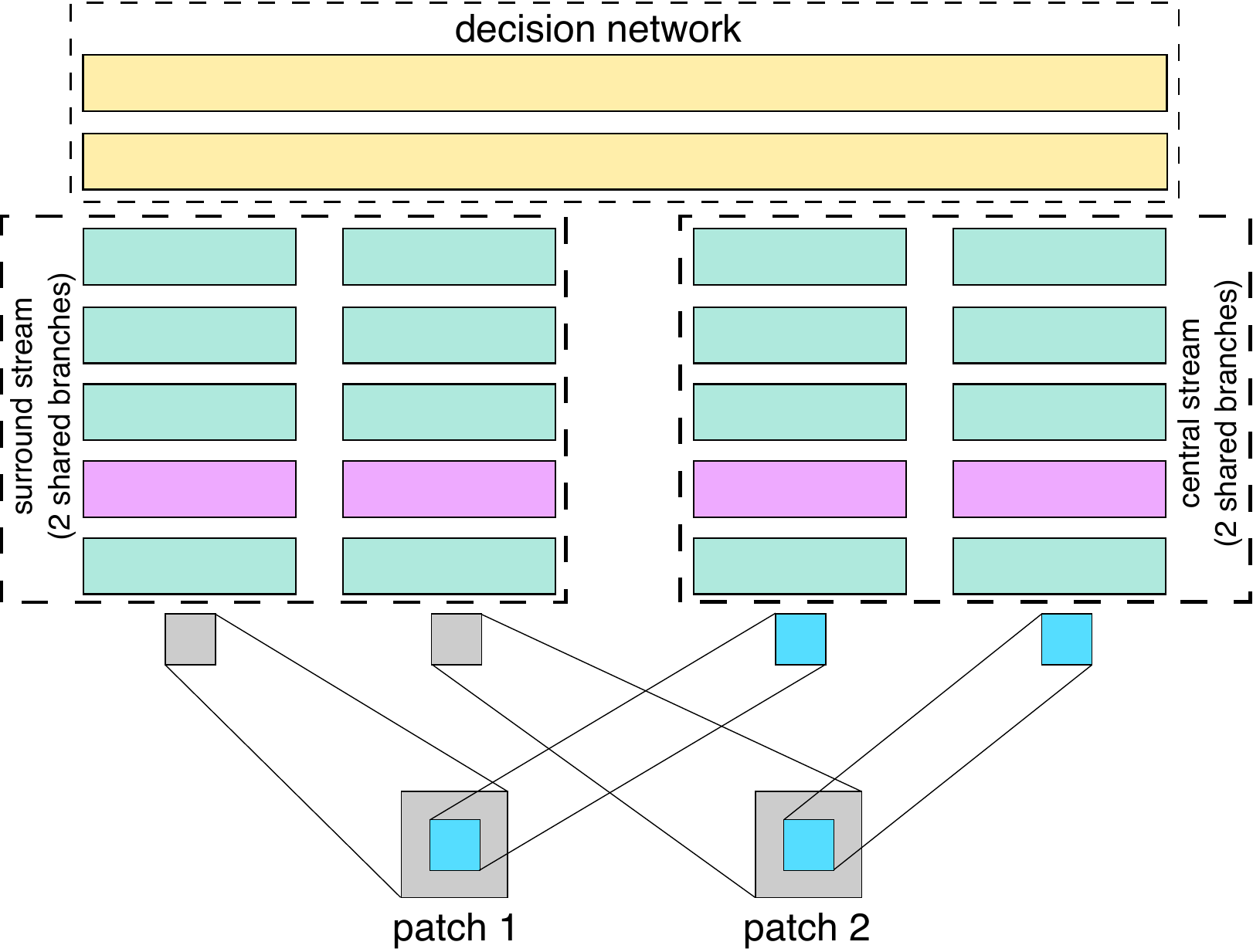}
\end{center}
\vspace{-8pt}
\caption{A central-surround two-stream network that uses a siamese-type architecture to process each stream. This results in 4 branches in total that are given as input to the top decision layer (the two branches in each stream are shared in this case).}
\vspace{-12pt}
\small
\label{fig:2stream}
\end{figure}
\fi

\textbf{Central-surround two-stream network.} As its name suggests, the proposed architecture consists of two separate streams, central and surround, which enable a processing in the spatial domain that takes place  over two different resolutions. More specifically, the central high-resolution stream receives as input two  $32\times 32$ patches that are generetad by cropping (at the original resolution) the central $32\times 32$ part of each input $64\times 64$ patch.
Furthermore, the surround low-resolution stream receives as input two   $32\times 32$ patches, which  are generated by downsampling
at half the original pair of input patches. The resulting two streams can then be processed by  using any of the basic architectures described in section \ref{sec:basic_models} (see Fig.~\ref{fig:2stream} for an example that uses a siamese architecture for each stream). 

One reason to make use of such a two-stream architecture is because multi-resolution information is  known to be important in improving the performance of image matching. Furthermore, by considering  the central part of a patch twice
(\ie, in both the high-resolution
and  low-resolution streams) we implicitly put more focus on the pixels closer to the center of a patch and less focus on the pixels in the periphery, which can also help for improving the precision of matching
(essentially, since  pooling is applied to the downsampled image,   pixels in the periphery are allowed to have more variance during matching). Note that  the total input dimenionality is  reduced by a factor of two in this case.
As a result, training  proceeds  faster, which is also one other practical advantage. 

\ifdefined\USEIMAGES
\begin{figure}
\small
\begin{center}
\includegraphics[scale=0.5]{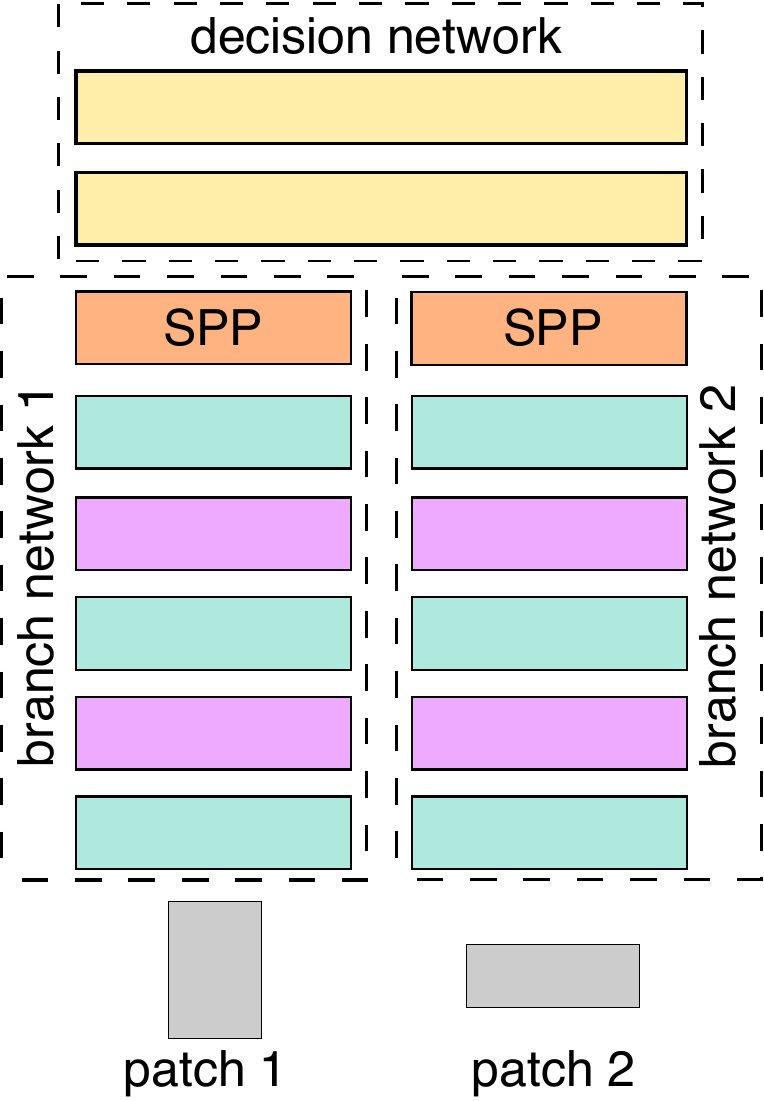}
\end{center}
\vspace{-5pt}
\caption{SPP network for a siamese architecture:  SPP layers (orange) are inserted immediately after the 2 branches of the network so that  the top decision layer has an input of  fixed dimensionality for any size of the input patches.}
\small
\label{fig:spp}
\vspace{-15pt}
\end{figure}
\fi
\textbf{Spatial pyramid pooling (SPP) network for comparing patches.}\label{sec:spp} Up to this point we have been assuming that the network requires the input patches to have a fixed size of $64\times 64$. This requirement comes from the fact that the output of the last convolutional layer of the network needs to have a predefined dimensionality. Therefore, when we need to compare patches of arbitrary sizes, this means that we first have to resize them to the above spatial dimensions. However, if we look at the example of descriptors like SIFT, for instance, we can see that another possible way to deal with patches of arbitrary sizes is via   adjusting  the size of the spatial pooling regions to be proportional to the size of the input patch so that we can still maintain the required fixed output dimensionality for the last convolutional layer without deteriorating the resolution of the input patches. %

This is also the idea behind the recently proposed SPP-net architecture \cite{spp}, which essentially amounts to inserting a spatial pyramid pooling layer between the convolutional layers and  the fully-connected layers of the network. Such a layer
aggregates the features of the last convolutional layer through spatial pooling, where the size of the pooling regions is dependent on the size of the input. Inspired by this, we propose to also consider adapting the network models of section \ref{sec:basic_models} according to the above SPP-architecture. This can be easily achieved for all the considered  models (\eg, see Fig.~\ref{fig:spp} for an example with a siamese model).

\section{Learning}

\textbf{Optimization.} We train all models in strongly supervised manner. We use  a hinge-based loss term and squared $l_2$-norm regularization that leads to the following learning objective function 
\begin{equation}\label{eq:tr_obj1}
\min_{w}\frac{\lambda}{2}\|w\|_2 + \sum_{i=1}^N\max(0, 1 - y_io^{net}_i)\enspace,
\end{equation}
where $w$ are the weights of the neural network, $o^{net}_i$ is the network output  for the $i$-th training sample, and $y_i \in \{-1,1\}$ the corresponding label (with $-1$ and $1$ denoting a non-matching and a matching  pair, respectively).

ASGD with constant learning rate 1.0, 
momentum 0.9 and weight decay $\lambda=0.0005$ is used to train the models. Training is done in 
mini-batches of size 128. Weights are initialised randomly and  all
 models are trained from scratch.

\textbf{Data Augmentation and preprocessing.} To combat overfitting we augment 
training data by flipping both patches in pairs horizontally and vertically and 
rotating to 90, 180, 270 degrees. As we don't notice overfitting while training 
in such manner we train models for a certain number of iterations, usually for 
2 days, and then test performance on test set.

Training dataset size allows us to store all the images directly in GPU memory and 
very efficiently retrieve patch pairs during training. Images are augmented 
"on-the fly". We use Titan GPU in Torch %
\cite{collobert:2011c} and
convolution routines  are taken from Nvidia cuDNN library \cite{cudnn}.
Our siamese descriptors on GPU are just 2 times slower than computing SIFT descriptors on CPU
and 2 times faster than Imagenet descriptors on GPU according to \cite{comparison}.

\section{Experiments}

We applied our models to a variety of problems and datasets. In the following we report results, and also provide comparisons with the state-of-the-art.

\subsection{Local image patches benchmark}
For a first evaluation of our models, we used the standard benchmark dataset from \cite{BHW10} that consists of three subsets, Yosemite, Notre Dame, and Liberty, each of 
which contains more than 450,000 image patches (64 x 64 pixels) sampled around 
Difference of Gaussians feature points. The patches are  scale and  orientation normalized. 
Each of the subsets was generated using actual 3D correspondences obtained via multi-view stereo depth maps. 
These maps were used to produce 500,000 ground-truth feature 
pairs for each dataset, with equal number of positive (correct) and negative (incorrect) matches.

For evaluating  our models, we use the evaluation
protocol of \cite{Brown11} and generate ROC curves by thresholding the distance between 
feature pairs in the descriptor space. We report the false positive rate at 
95\% recall (FPR95) on each of the six combinations of training and test sets, 
as well as the mean across all combinations. We also report the mean, denoted as $\mathrm{mean}(1,4)$, for only those 4 combinations that were used in \cite{Boix13CVPR}, \cite{BHW10}  (in which case training takes place on Yosemite or Notre Dame, but not 
Liberty).

\begin{table*}[t]
\footnotesize
\begin{center}
\begin{tabular}{rrcccccccccc}
\hline
    Train  &  Test     & 2ch-2stream & 2ch-deep  & 2ch   & siam & siam-$l_2$ & pseudo-siam & pseudo-siam-$l_2$ & siam-2stream & siam-2stream-$l_2$ & \cite{Simonyan14} \\ \hline
Yos   & ND    & \textbf{2.11} & 2.52 & 3.05  & 5.75  & 8.38  & 5.44  & 8.95  & 5.29 & 5.58 & 6.82 \\
Yos   & Lib   & \textbf{7.2} & 7.4 & 8.59  & 13.48 & 17.25 & 10.35 & 18.37 & 11.51 & 12.84 &14.58 \\
ND    & Yos   & \textbf{4.1} & 4.38 & 6.04  & 13.23 & 15.89 & 12.64 & 15.62 & 10.44 & 13.02 & 10.08 \\
ND    & Lib   & 4.85 & \textbf{4.55} & 6.05  & 8.77  & 13.24 & 12.87 & 16.58 & 6.45 & 8.79 & 12.42 \\
Lib   & Yos   & 5 & \textbf{4.75} & 7     & 14.89 & 19.91 & 12.5  & 17.83 & 9.02  & 13.24 & 11.18 \\
Lib   & ND    & \textbf{1.9} & 2.01 & 3.03  & 4.33  & 6.01  & 3.93  & 6.58  & 3.05 & 4.54 & 7.22 \\
\hline
\multicolumn{2}{c}{mean} & \textbf{4.19} & 4.27 & 5.63  & 10.07 & 13.45 & 9.62  & 13.99 & 7.63 & 9.67 & 10.38 \\
\hline
\multicolumn{2}{c}{mean(1,4)} & \textbf{4.56} & 4.71 & 5.93  & 10.31 & 13.69 & 10.33 & 14.88 & 8.42 & 10.06 & 10.98 \\
\hline
\end{tabular}
\end{center}
\footnotesize
\caption{Performance of several models on the ``local image patches'' benchmark.
 The models architecture  is as follows:
 (i) $2\mathrm{ch}$-$2\mathrm{stream}$ consists of two branches $\mathrm{C}(95,5,1)$-$\mathrm{ReLU}$-$\mathrm{P}(2,2)$-$\mathrm{C}(96,3,1)$-$\mathrm{ReLU}$-$\mathrm{P}(2,2)$-$\mathrm{C}(192,3,1)$-$\mathrm{ReLU}$-$\mathrm{C}(192,3,1)$-$\mathrm{ReLU}$, one for central and one for surround parts, followed by $\mathrm{F}(768)$-$\mathrm{ReLU}$-$\mathrm{F}(1)$
(ii) $2\mathrm{ch}$-$\mathrm{deep}=\mathrm{C}(96,4,3)$-$\mathrm{Stack}(96)$-$\mathrm{P}(2,2)$-$\mathrm{Stack}(192)$-$\mathrm{F}(1)$, where $\mathrm{Stack}(n)=\mathrm{C}(n,3,1)$-$\mathrm{ReLU}$-$\mathrm{C}(n,3,1)$-$\mathrm{ReLU}$-$\mathrm{C}(n,3,1)$-$\mathrm{ReLU}$.
(iii) $2\mathrm{ch}=\mathrm{C}(96,7,3)$-$\mathrm{ReLU}$-$\mathrm{P}(2,2)$-$\mathrm{C}(192,5,1)$-$\mathrm{ReLU}$-$\mathrm{P}(2,2)$-$\mathrm{C}(256,3,1)$-$\mathrm{ReLU}$-$\mathrm{F}(256)$-$\mathrm{ReLU}$-$\mathrm{F}(1)$
(iv) $\mathrm{siam}$ has two branches $\mathrm{C}(96,7,3)$-$\mathrm{ReLU}$-$\mathrm{P}(2,2)$-$\mathrm{C}(192,5,1)$-$\mathrm{ReLU}$-$\mathrm{P}(2,2)$-$\mathrm{C}(256,3,1)$-$\mathrm{ReLU}$ and decision layer $\mathrm{F}(512)$-$\mathrm{ReLU}$-$\mathrm{F}(1)$
(v) $\mathrm{siam}$-$l_2$ reduces to a single branch of $\mathrm{siam}$ 
(vi) $\mathrm{pseudo}$-$\mathrm{siam}$ is uncoupled version of $\mathrm{siam}$
(vii) $\mathrm{pseudo}$-$\mathrm{siam}$-$l_2$ reduces to a single branch of $\mathrm{pseudo}$-$\mathrm{siam}$ 
(viii) $\mathrm{siam}$-$\mathrm{2stream}$ has 4 branches $\mathrm{C}(96,4,2)$-$\mathrm{ReLU}$-$\mathrm{P}(2,2)$-$\mathrm{C}(192,3,1)$-$\mathrm{ReLU}$-$\mathrm{C}(256,3,1)$-$\mathrm{ReLU}$-$\mathrm{C}(256,3,1)$-$\mathrm{ReLU}$ (coupled in pairs for central and surround streams), and decision layer $\mathrm{F}(512)$-$\mathrm{ReLU}$-$\mathrm{F}(1)$
(ix) $\mathrm{siam}$-$\mathrm{2stream}$-$l_2$ consists of one central and one surround branch of $\mathrm{siam}$-$\mathrm{2stream}$. 
 The shorthand notation used was the following: $\mathrm{C}(n,k,s)$ is a convolutional layer with $n$ filters of spatial size $k\times k$ applied with stride $s$, $\mathrm{P}(k,s)$ is a max-pooling layer of size $k\times k$ applied with stride $s$, and $\mathrm{F}(n)$ denotes a fully connected linear layer with $n$ output units.}
\label{maintable}
\end{table*}

Table~\ref{maintable} reports the performance of several models, and also details their architecture (we  have also experimented with smaller kernels, less max-pooling layers, as well as adding normalisations, without noticing any significant improvement in performance).
We briefly summarize some of the conclusions that can be drawn from this table. A first important conclusion  is that 
  2-channel-based architectures (\eg, \texttt{2ch}, \texttt{2ch-deep}, \texttt{2ch-2stream})   exhibit clearly the best performance among all models. This  is something that indicates that  it is important to \emph{jointly} use information from both patches {right from the \emph{first} layer} of the network.

\texttt{2ch-2stram} network  was the top-performing network on this dataset, with  \texttt{2ch-deep} following closely (this verifies the importance of multi-resolution information during matching and that also increasing the network depth helps). In fact,  \texttt{2ch-2stream} managed  to outperform the previous state-of-the-art  by a large margin, achieving $2.45$ times better score than    \cite{Simonyan14}! The difference with   SIFT was even larger, with our model giving  $6.65$ times better score in this case (SIFT score  on \texttt{mean(1,4)} was $31.2$ according to \cite{BHW10}).

Regarding siamese-based architectures, these too manage to achieve  better performance than  existing state-of-the-art systems. This is quite interesting because, \eg, none of these siamese networks tries to learn the shape, size or placement of the pooling regions (like, \eg, \cite{Simonyan14, BHW10} do), but instead utilizes just standard max-pooling layers.
Among the siamese models, the two-stream network (\texttt{siam-2stream}) had the best performance,  
verifying once more the importance of multi-resolution information  when it comes to comparing image patches. Furthermore,  the pseudo-siamese network (\texttt{pseudo-siam}) was better than the corresponding siamese one (\texttt{siam}).

We also conducted additional experiments, in which we tested the performance of siamese models when their top decision layer is replaced with the $l_2$ Euclidean distance of the two convolutional descriptors produced by the two branches of the network (denoted with the suffix $l_2$ in the name). In this case, prior to applying the Euclidean distance, the descriptors are $l_2$-normalized (we also tested $l_1$ %
normalization). 
For pseudo-siamese only one branch was used
to extract descriptors. As expected, in this case the two-stream network (\texttt{siam-2stream-}$l_2$) computes better distances  than the siamese network (\texttt{siam-}$l_2$), which, in turn, computes better distances than the pseudo-siamese model (\texttt{pseudo-siam-}$l_2$).  In fact,  the  \texttt{siam-2stream-}$l_2$ network manages to outperform even the previous state-of-the-art descriptor \cite{Simonyan14}, which is quite surprising given that these siamese models have never been trained  using $l_2$ distances.
 
 \ifdefined\USEIMAGES
\begin{figure*}[t]
\begin{center}
\includegraphics[scale=0.45]{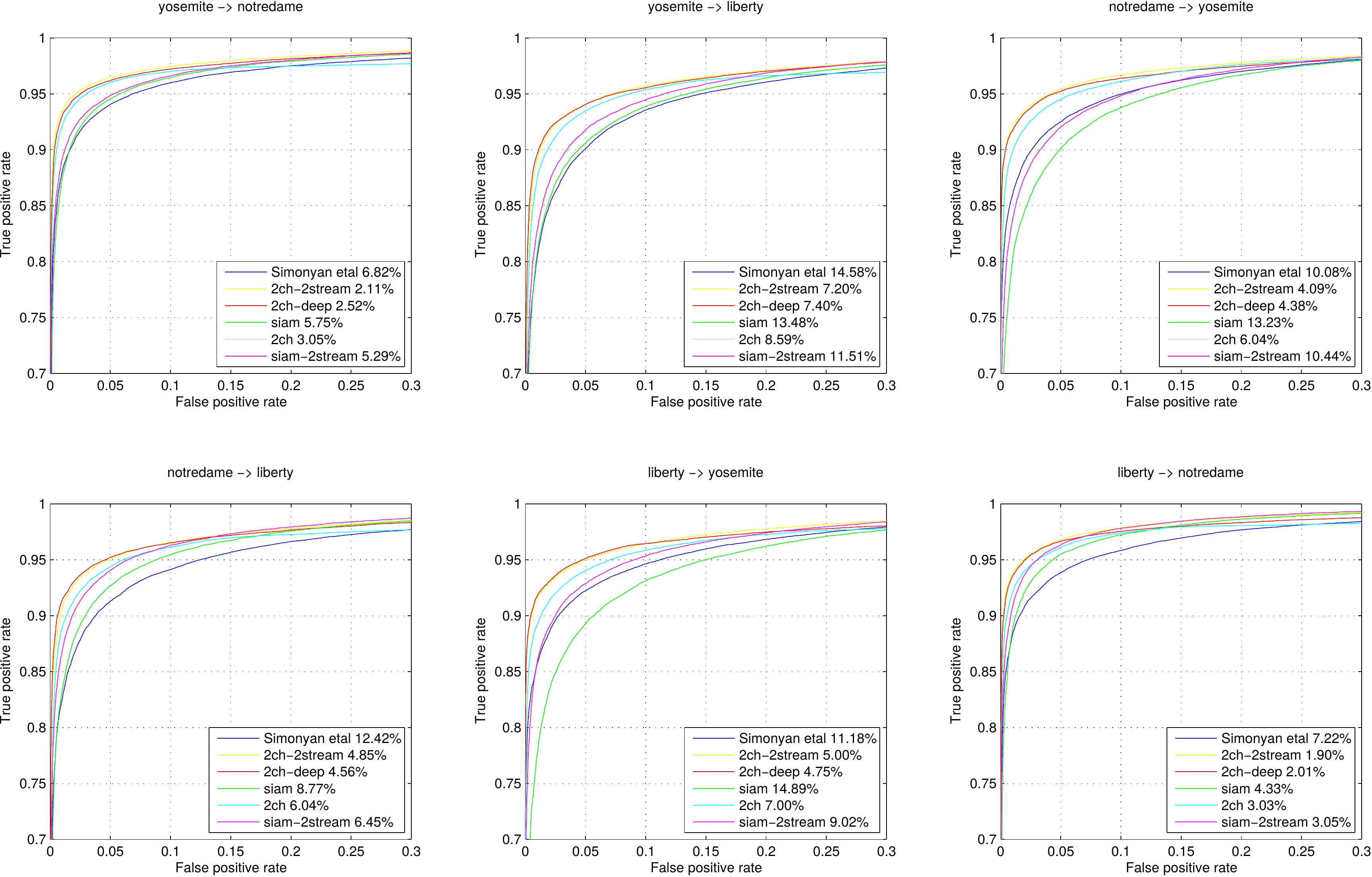}
\end{center}
\vspace{-6pt}
\caption{ROC curves for various models (including the state-of-the-art descriptor \cite{Simonyan14}) on the local image patches benchmark. Numbers in the legends are corresponding FPR95 values}
\vspace{-3pt}
\label{fig:standard_benchmark_plots}
\end{figure*}
\fi

\begin{table}

\begin{center}
\small
\begin{tabular}{lcccccc}
\hline
 & $\mathrm{conv3}_{(3456)}$  &   $\mathrm{conv4}_{(3456)}$  & $\mathrm{conv5}_{(2304)}$ \\ 
\hline
Notredame     & 12.22  & 9.64  & 19.384 \\
Liberty            & 16.25  & 14.26  & 21.592 \\
Yosemite        & 33.25  & 30.22  & 43.262 \\ \hline
mean  & 20.57  &17.98  &  28.08 \\
\hline
\end{tabular}
\small
\end{center}
\caption{FPR95 for imagenet-trained features (dimensionality of each feature appears as subscript).}
\label{standard_benchmark_imagenet}
\vspace{-10pt}
\end{table}

For a more detailed comparison of the various models, we  provide the corresponding ROC curves in Fig.~\ref{fig:standard_benchmark_plots}. Furthermore,  we show in Table~\ref{standard_benchmark_imagenet} the performance of imagenet-trained CNN features (these were $l_2$-normalized to improve results). Among these, \texttt{conv4} gives the best FPR95 score, which is equal to $17.98$. This makes it better than SIFT but still much worse than our models. 

\ifdefined\USEIMAGES
\begin{figure}[h]
\centering
\subfigure[]{
        \includegraphics[scale=0.25]{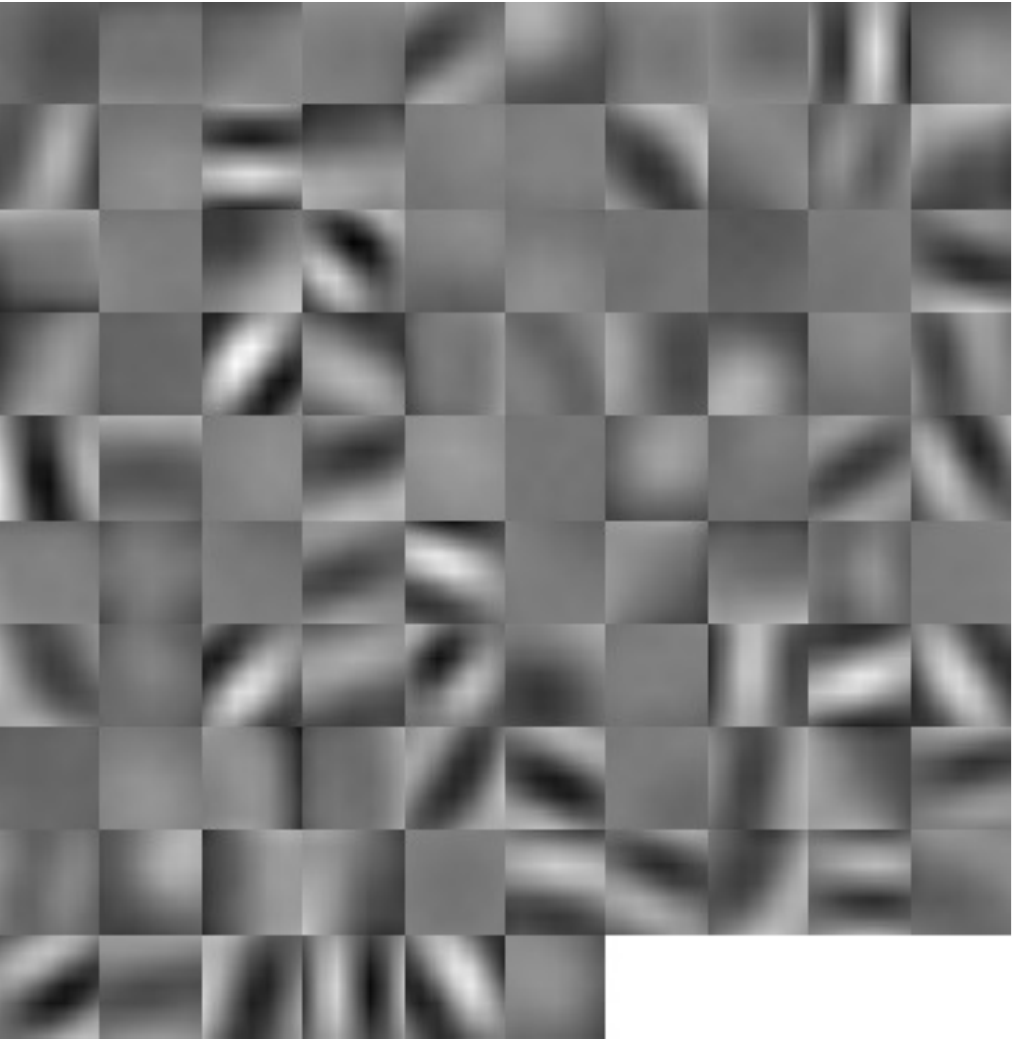}
        \label{fig:siamese_filters}
}\hspace{1cm}
\subfigure[]{
        \includegraphics[scale=0.3]{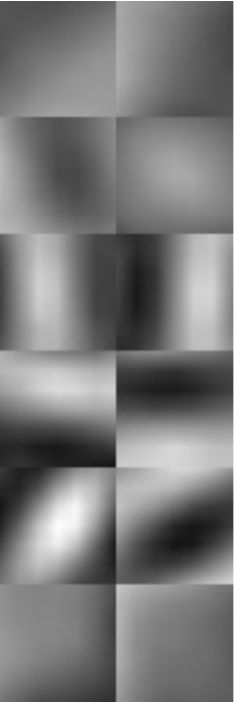}
        \label{fig:2ch_filters}
}
\caption{(a) Filters of the first convolutional layer of \texttt{siam} network. (b) Rows correspond to  first layer filters from \texttt{2ch} network (only a subset shown), depicting left and right  part of each filter.}
\end{figure}
\vspace{-15pt}
\fi

\ifdefined\USEIMAGES
\begin{figure}[h]
\centering
\subfigure[true positives]{\includegraphics[scale=0.22]{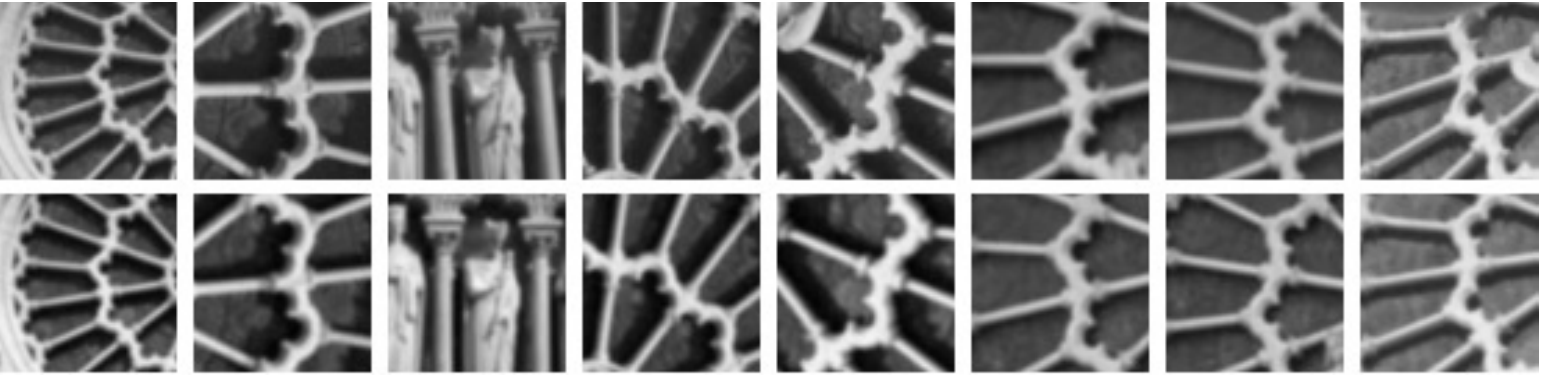}}
\subfigure[false negatives]{\includegraphics[scale=0.22]{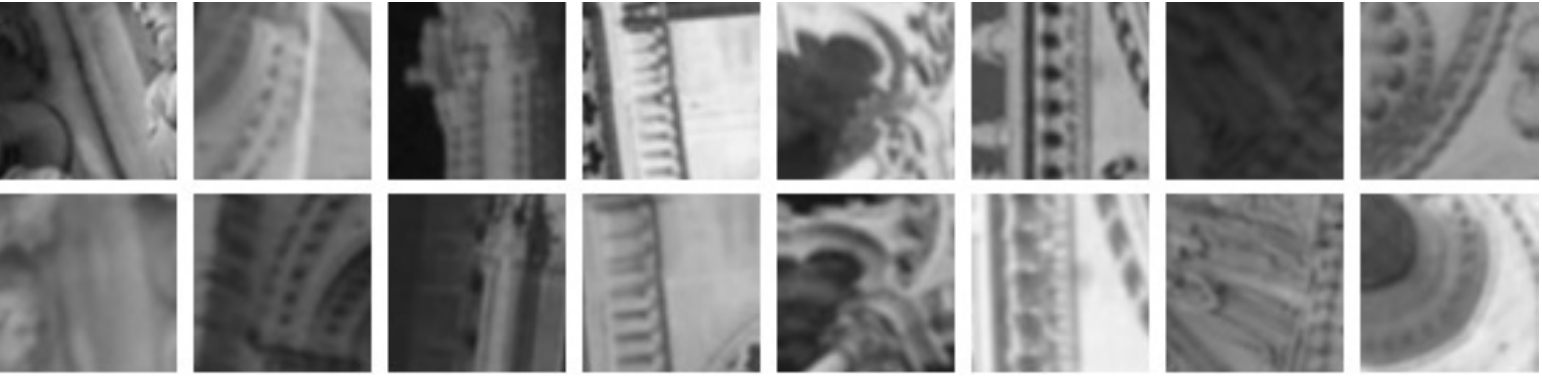}}
\subfigure[true negatives]{\includegraphics[scale=0.22]{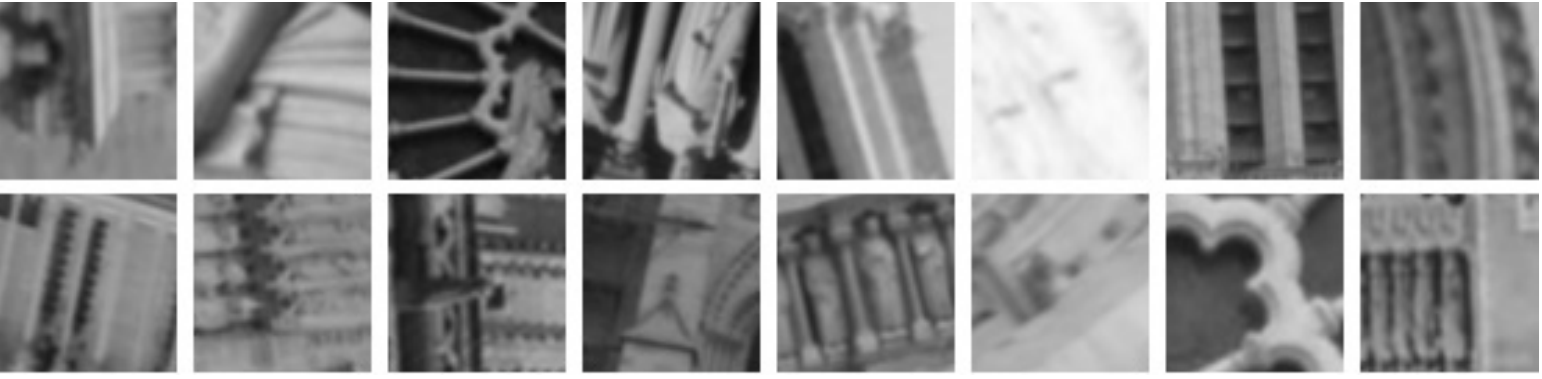}}
\subfigure[false positives]{\includegraphics[scale=0.22]{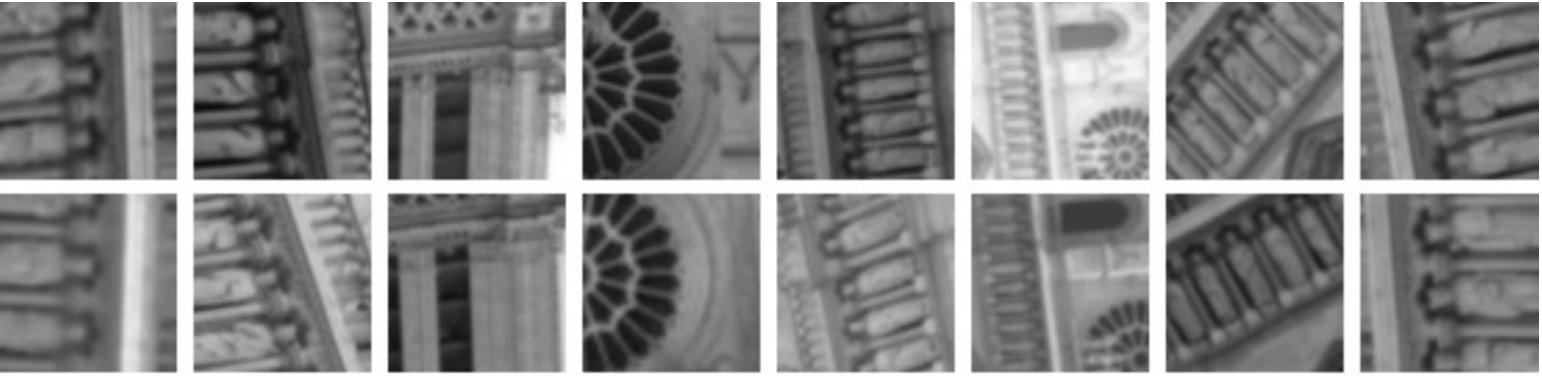}}
\caption{Top-ranking false and true matches by \texttt{2ch-deep}.}
\label{fig:false_true_matches}
\vspace{-5pt}
\end{figure}
\fi

Fig.~\ref{fig:siamese_filters} displays the filters of the first convolutional layer learnt by the siamese network. Furthermore, Fig.~\ref{fig:2ch_filters} shows the left and right parts for a subset of the first layer filters of the 2-channel network \texttt{2ch}. It is worth mentioning that corresponding left and right parts look like being negative to each other, which basically means that the network has learned to  compute differences of features between the two patches 
(note, though, that not all first layer filters of \texttt{2ch}  look like this). Last, we show in Fig.~\ref{fig:false_true_matches} some top ranking false and  correct matches as computed by the \texttt{2ch-deep} network. We observe that false matches could be easily mistaken even by a human (notice, for instance, how similar the two patches in false positive examples look like).

For the rest of the experiments, we note that we use models trained on the Liberty dataset.

\subsection{Wide baseline stereo evaluation}

For this evaluation we chose the dataset by Strecha \etal\cite{StrechaDataset}, which contains several image sequences with ground truth homographies and 
laser-scanned depthmaps. We used ``fountain'' and ``herzjesu'' sequences to produce 6 and 5
rectified stereo pairs respectively. Baselines in both sequences we chose are 
increasing with each image making matching more difficult. Our goal was to show that 
a photometric cost computed with neural network competes favorably against costs produced
by a state-ot-the-art hand-crafted feature descriptor, so we chose to compare with DAISY \cite{Tola08}.

Since our focus was not on efficiency, we used an unoptimized pipeline for computing the photometric costs. More specifically,  for 2-channel networks we used a brute-force approach, where we extract patches on corresponding epipolar lines with subpixel estimation, construct
batches (containing a patch from the left image $I_1$ and all patches on the corresponding epipolar 
line from the right image $I_2$) and compute network outputs, resulting in the cost:
\begin{equation}
  C(\mathbf{p},d) =  - o^{net}(I_1(\mathbf{p}), I_2(\mathbf{p} + d))
  \label{eq:cnet}
\end{equation}
Here,  $I(\mathbf{p})$ denotes a neighbourhood intensity matrix around a pixel $\mathbf{p}$,
$o^{net}(P_1, P_2)$ is the output of the neural network given a pair of patches
$P_1$ and $P_2$, and $d$ is the  distance between points on epipolar line.

For siamese-type networks, %
we compute descriptors for each pixel
in both images once and then match them with decision top layer or $l_2$ distance. In the first case
the formula for photometric cost is the following:
\begin{equation}
  C(\mathbf{p},d) = - o^{top}(D_1(I_1(\mathbf{p})), D_2(I_2(\mathbf{p} + d)))
\end{equation}
where $o^{top}$ is output of the top decision layer, and $D_1$, $D_2$ are outputs of branches of the siamese
or pseudo-siamese network, i.e.\ descriptors (in case of siamese network $D_1 = D_2$). For $l_2$ matching, it holds:
\begin{equation}
  C(\mathbf{p},d) = \|D_1(I_1(\mathbf{p})) - D_2(I_2(\mathbf{p} + d))\|_2
\end{equation}
It is worth noting that all costs above can be computed a lot more efficiently  using  speed optimizations similar with \cite{Zbontar}. This essentially means treating all fully connected layers as $1\times 1$ convolutions, computing branches of siamese network  only once, and furthermore computing the outputs of these branches as well as the final outputs of the network at all locations using a number of forward passes on full images (e.g.,  for a 2-channel architecture  such an approach of computing the photometric costs  would only require  feeding the network with $s^2\cdot d_{\mathrm{max}}$ full 2-channel images of size equal to the input image pair, where $s$ is the stride at the first layer of the network and $d_{\mathrm{max}}$ is the maximum disparity).

Once computed, the  photometric costs are  subsequently used as unary terms in the following pairwise MRF energy  
\[
E(\{d_p\}) \!=\! \sum_pC(\mathbf{p},d_p) +\!\!\! \sum_{(p,q)\in \mathcal{E}}\!\!\!(\lambda_1 + \lambda_2e^{-\frac{\|{\nabla}I_1(\mathbf{p})\|_2}{\sigma^2}}) \cdot|d_p-d_q|\,,
\]
minimized using algorithm \cite{NIPS2014_IbyL}  based on FastPD \cite{komodakis_cvpr2007} (we set $\lambda_1\!=\!0.01$, $\lambda_2\!=\!0.2$, $\sigma\!=\!7$ and $\mathcal{E}$ is a 4-connected grid).

\ifdefined\USEIMAGES
\begin{figure*}
\includegraphics[width=0.98\linewidth]{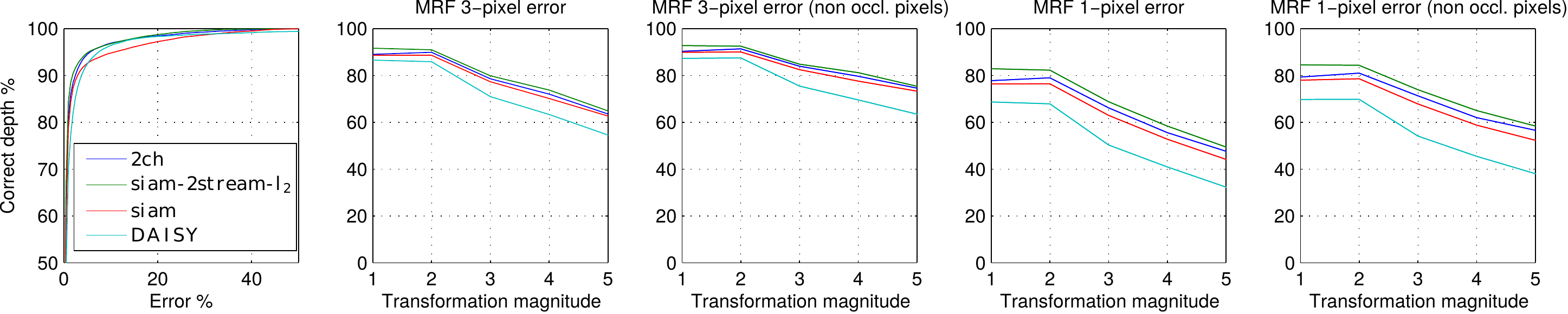}
\caption{Quantitative comparison for wide-baseline stereo on ``fountain'' dataset. (Leftmost plot) Distribution of deviations from ground truth, %
expressed as a fraction of scene's depth range. (Other plots) Distribution of errors for stereo pairs of increasing baseline (horizontal axis) both with and without taking into account  occluded pixels (error thresholds were set equal to 1 and 3 pixels in these plots -  maximum disparity is around 500 pixels). 
}
\label{fig:stereo_quantitative}
\end{figure*}
\fi
We  show in Fig.~\ref{fig:stereo_qualitative} some qualitative results in terms of computed depth maps  (with and without global optimization) for the ``fountain'' image set (results for ``herzjesu'' appear in  supp. material due to lack of space). Global
MRF optimization results   visually verify that photometric cost computed with neural network
is much more robust than with hand-crafted features, as well as the high quality of the depth maps produced by 2-channel architectures. Results without global optimization also show that the estimated  depth maps  contain much more fine details than DAISY. They may exhibit a very sparse set of errors  for the case of siamese-based networks,  but these errors can be very easily eliminated during global optimization.

Fig.~\ref{fig:stereo_quantitative} also shows a quantitative comparison, focusing in this case on siamese-based models as they are more efficient. The first plot of that figure shows (for a single stereo pair) the distribution of deviations from the ground truth across all range of error thresholds (expressed here as a fraction of the scene's depth range).
Furthermore, the other plots of the same figure summarize the corresponding distributions of errors for the six stereo pairs of increasing baseline (in this case we  also  show separately the error distributions when  only unoccluded pixels are taken into account). The error thresholds  were set to 3 and 5 pixels  in these plots (note that the maximum disparity is around 500 pixels in the largest baseline). 
As can be seen, all siamese models 
perform much better than DAISY across all error thresholds and all baseline distances 
(\eg, notice the difference in the curves of the corresponding plots).

\ifdefined\USEIMAGES
\begin{figure*}
  \vspace{-5pt}
\begin{center}
  \includegraphics[height=4cm]{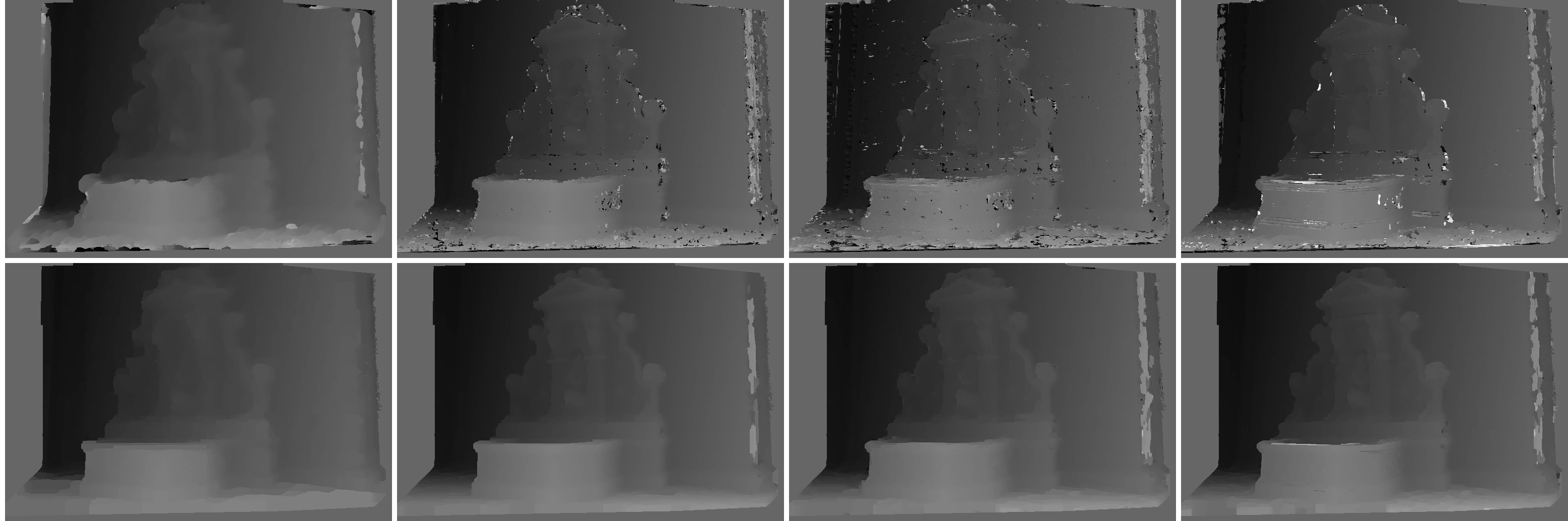}
\end{center}
\vspace{-5pt}
\caption{Wide baseline stereo evaluation. From left to right:
  DAISY, \texttt{siam-2stream-}$l_2$, \text{siam}, \texttt{2ch}. First row - ``winner takes all'' depthmaps,
  second row - depthmaps after MRF optimization.}
  \label{fig:stereo_qualitative}
\end{figure*}
\fi

\subsection{Local descriptors performance evaluation}

We also test our networks on Mikolajczyk dataset for
local descriptors evaluation \cite{MS05}. The dataset consists of 48 images in 6 sequences
with camera viewpoint changes, blur, compression, lighting changes and zoom
with gradually increasing amount of transfomation. There are known ground 
truth homographies between the first and each other image in sequence.

Testing technique is the same as in \cite{MS05}. 
Briefly, to test a pair of images, detectors are applied
to both images to extract keypoints. Following \cite{comparison}, we use 
MSER detector. 
The ellipses provided by detector are used
to exctract patches from input images. Ellipse size is magnified by a factor of 3
to include more context.
Then, depending on the  type of network, either descriptors, meaning outputs of siamese
or pseudo-siamese branches, are extracted, or all patch pairs are given to 2-channel
network to assign a score.

A quantitative comparison  on this dataset is shown for several models in Fig.~\ref{fig:local_descriptors}.
Here we also test the CNN network  \texttt{siam-SPP-}$l_2$, which is an SPP-based siamese architecture (note that \texttt{siam-SPP} is same as \texttt{siam} but with the addition of two SPP layers - see also Fig.~\ref{fig:spp}). We used an inserted SPP layer that had a spatial dimension of $4\times 4$. As can be seen, this  provides a 
big boost in matching performance,  suggesting the great utility of such an architecture  
when comparing image patches. 
Regarding the rest  of the  models, the observed results in Fig.~\ref{fig:local_descriptors} reconfirm the conclusions already drawn  from  previous experiments. We simply note again the very good performance of \texttt{siam-2stream-}$l_2$, which (although not trained with $l_2$ distances) is able to significantly outperform SIFT and to also match the performance of imagenet-trained features (using, though, a much lower dimensionality of $512$).

\section{Conclusions}
In this paper we showed how to  learn  directly from raw image pixels a  general similarity function for patches, which is encoded in the form of a CNN model.  To that end, we studied  several neural network architecures  that are specifically  adapted to this task, and showed that they exhibit extremely good performance, significantly outperforming the state-of-the-art on several problems and benchmark datasets.

\ifdefined\USEIMAGES
\begin{figure}
\vspace{-5pt}
  \includegraphics[width=1\linewidth]{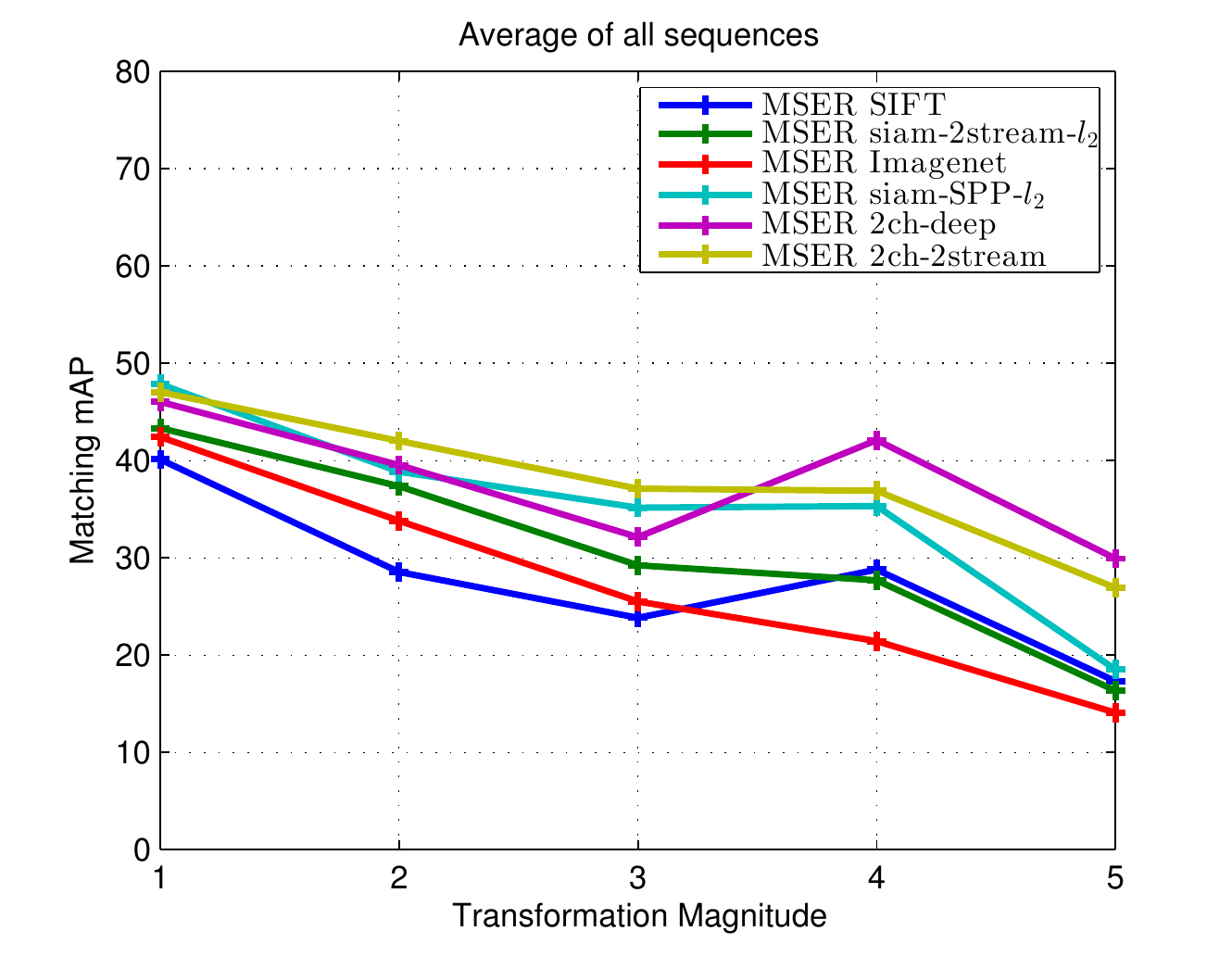}
\caption{Evaluation on the Mikolajczyk dataset \cite{MS05} showing the mean average precision (mAP) averaged over all types of transformations in the dataset (as usual, the mAP score measures the area under the precision-recall curve).
More detailed plots are provided  in the supplemental material due to lack of space.}
\label{fig:local_descriptors}
\vspace{-15pt}
\end{figure}
\fi

Among these architectures, we note that 2-channel-based ones were clearly the superior in terms of results. It  is, therefore, worth investigating how to further accelerate the evaluation of these networks in the future. Regarding siamese-based architectures, 2-stream multi-resolution models turned out to be extremely strong,  providing always a significant boost in performance and verifying the importance of multi-resolution information when comparing patches. The same conclusion applies to SPP-based siamese networks, which  also consistently  improved the quality of  results\footnote{In fact, SPP performance can improve even further, as no multiple aspect ratio patches were used during the training of SPP models (such patches appear only at test time).}. 

Last, we should note that simply the use of a larger training set can potentially   benefit and improve the overall performance of our approach  even further (as the training set that was used in the  present experiments can actually be considered rather small by today's standards).

{\small
\bibliographystyle{ieee}
\bibliography{cvpr2015}
}

\end{document}